\documentclass[conference]{IEEEtran}
\IEEEoverridecommandlockouts
\usepackage{cite}
\usepackage{amsmath,amssymb,amsfonts}
\usepackage{algorithmic}
\usepackage{graphicx}
\usepackage{textcomp}
\usepackage{xcolor}
\usepackage{enumitem}
\usepackage{balance}

\def\BibTeX{{\rm B\kern-.05em{\sc i\kern-.025em b}\kern-.08em
    T\kern-.1667em\lower.7ex\hbox{E}\kern-.125emX}}
\begin{document}

\title{A Secure and Trustworthy Network Architecture for Federated Learning Healthcare Applications
\thanks{Funded by the EU in the call HORIZON-HLTH-2022-STAYHLTH-01-twoStage under grant agreement No 101080564.}
}

\author{\IEEEauthorblockN{{Antonio Boiano}$^1$, Marco Di Gennaro$^1$, Luca Barbieri$^1$, Michele Carminati$^1$, Monica Nicoli$^1$, Alessandro Redondi$^1$ \\Stefano Savazzi$^2$, Albert Sund Aillet$^3$, Diogo Reis Santos$^3$, Luigi Serio$^3$}
\IEEEauthorblockA{\textit{$^1$DEIB, Politecnico di Milano, Milan, Italy} \\
\textit{$^2$IEIIT, Consiglio Nazionale delle Ricerche (CNR), Milan, Italy}\\
\textit{$^3$CERN, Geneva, Switzerland}
}
}

\maketitle

\begin{abstract}
Federated Learning (FL) has emerged as a promising approach for privacy-preserving machine learning, particularly in sensitive domains such as healthcare. In this context, the TRUSTroke project aims to leverage FL to assist clinicians in ischemic stroke prediction. This paper provides an overview of the TRUSTroke FL network infrastructure. The proposed architecture adopts a client-server model with a central Parameter Server (PS). We introduce a Docker-based design for the client nodes, offering a flexible solution for implementing FL processes in clinical settings. The impact of different communication protocols (HTTP or MQTT) on FL network operation is analyzed, with MQTT selected for its suitability in FL scenarios. A control plane to support the main operations required by FL processes is also proposed. The paper concludes with an analysis of security aspects of the FL architecture, addressing potential threats and proposing mitigation strategies to increase the trustworthiness level.
\end{abstract}


\section{Introduction}\label{sec:intro}

FL is a decentralized machine learning approach where a model is trained across multiple edge nodes holding local data samples, without exchanging them. Local data remain on the nodes where they originate, and only model updates are shared, preserving the ownership and privacy of individual data samples. Compared to classical centralized learning, where all data are transferred to a central server for model training, FL offers clear advantages in terms of privacy, communication efficiency, scalability, and robustness compared to centralized learning, particularly in scenarios where data privacy is a concern or where data is originally distributed across multiple nodes \cite{abdulrahman2020survey,lim2020federated,mothukuri2021survey}. FL has become particularly appealing in the context of medical and healthcare data, which is generally subject to very strict privacy regulations. In this scenario, the TRUSTroke project \cite{trustroke} proposes a novel AI-based platform to assist clinicians, patients and caregivers in the management of acute and chronic phases of ischemic stroke. To this purpose, the project has the ambitious goal to design, build and maintain a FL infrastructure that will enable multiple hospitals and clinical sites to collaborate in training AI models able to predict several variables related to stroke (e.g., clinical severity at discharge, stroke recurrence, etc.) without sharing their data, hence without compromising privacy. The FL platform envisioned by the TRUSTroke project focuses particularly on trustworthiness, robustness and privacy-preservation in all its building blocks, in line with the recent EU Artificial Intelligence (AI) Act requirements. Such requirements hold not only for the AI algorithms that will be developed for stroke prediction, but extend also to any system-level component needed to run the FL platform, including networking and communication aspects. 
In this regard, this paper proposes an efficient, robust and secure network architecture, together with the related communication protocols, to support trustworthy FL processes. In detail, the contributions of this paper are the following:

\begin{enumerate}[leftmargin=*]
    \item \textit{Network architecture:} we propose in Section \ref{sec:network} a client-server architecture for the FL process, where a central Parameter Server (PS) aggregates models trained locally at different clinical clients. Both the PS and the clients are designed in a containerized fashion to provide a portable, efficient, secure and trusty operation. 
    
    \item \textit{Communication protocols:} we discuss the implications of using two different communication paradigms, i.e., Request/Response or Publish/Subscribe to support the operation of the FL platform. This is done in Section \ref{sec:comm} by comparing two widely popular protocols of different types, HTTP and MQTT: the latter is selected as reference communication protocol for the proposed FL platform due to a series of features which better fit with the FL scenario.
    
    \item \textit{Control and data plane design:} Section \ref{sec:plane} provides an analysis of the requirements of a FL platform in terms of fundamental operations the Parameter Server and the clinical clients need to perform. Based on such an analysis, MQTT-based control and data plane are proposed to support each FL operation efficiently.

    \item \textit{Security considerations:} additionally, Section \ref{sec:security} provides a summary of the security threats and risks associated with each component of the proposed architecture, as well as mitigation techniques and recommendations to be compiled in order to increase the security and trust level.

\end{enumerate}
Finally, Section \ref{sec:concl} concludes the paper and discusses future research directions.
\section{TRUSTroke Network Architecture}\label{sec:network}
\begin{figure}[t]
        \includegraphics[width=\columnwidth]{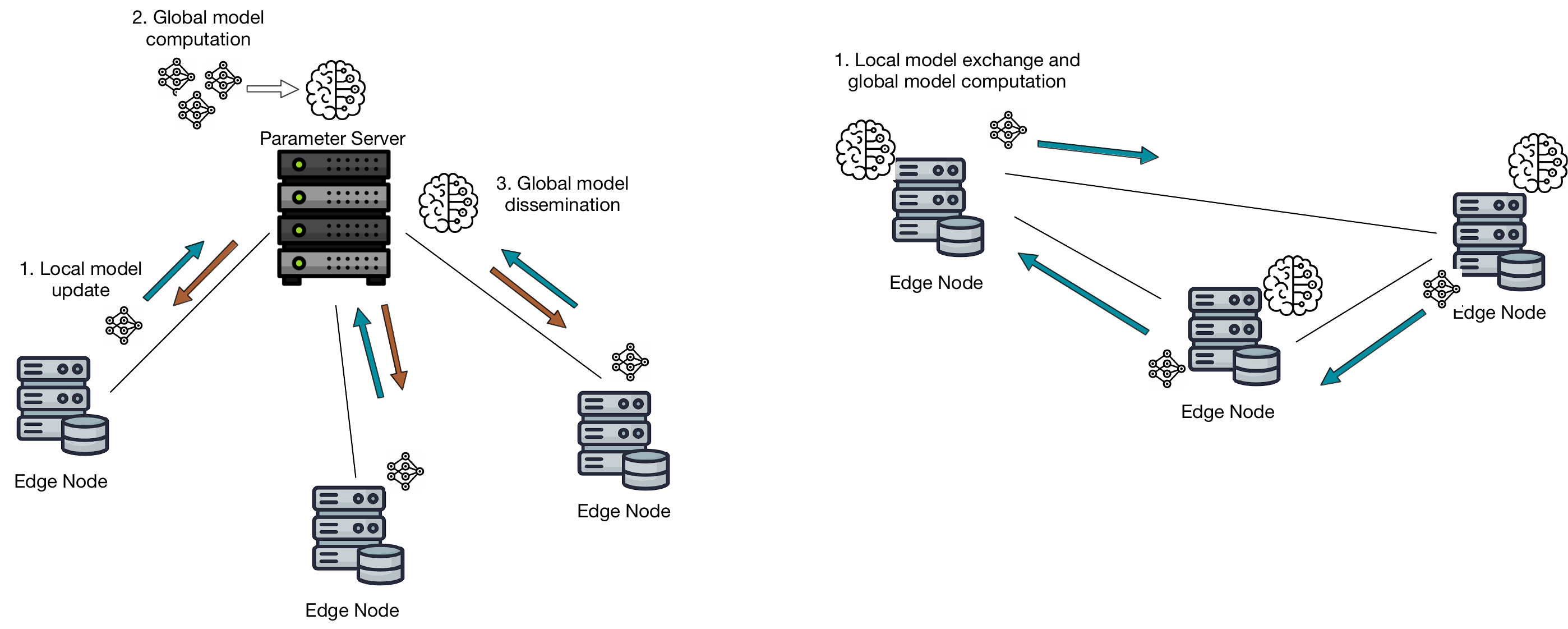}
        \caption{Centralized (left) vs Distributed (right) FL architectures}
        \label{fig:centralized_distributed}
\end{figure}

Two main communication architectures are typically utilized in FL scenarios: Client-Server (centralized) and Peer-to-Peer (decentralized) \cite{tedeschini2022decentralized}. In the centralized approach (Figure \ref{fig:centralized_distributed}, left), edge nodes exchange model parameters with a central PS, which aggregates these parameters to create a global model disseminated back to the nodes. In contrast, decentralized FL involves nodes exchanging parameters directly, without a central entity (Figure \ref{fig:centralized_distributed}, right). Decentralized FL offers advantages in communication efficiency and privacy but requires complex orchestration and may suffer from inconsistency among nodes. Hence, it is used with a massive number of nodes or when a trusted centralized server isn't available \cite{sun2022decentralized}.

The TRUSTroke project leverages the involvement of CERN among its partners and therefore opts for a centralized FL architecture. Indeed, CERN's powerful computing resources, stringent security measures, and well-known global collaboration network makes it an ideal candidate for hosting the PS. CERN's infrastructure ensures data privacy and security, providing a trusted environment for handling sensitive model updates. Therefore, the project implements a centralized FL architecture, with CERN as the PS and clinical partners as the edge nodes, as illustrated in Figure \ref{fig:network_architecture}.

\subsection{Parameter server}

\begin{figure*}[t]
        \centering
        \includegraphics[width=0.9\textwidth]{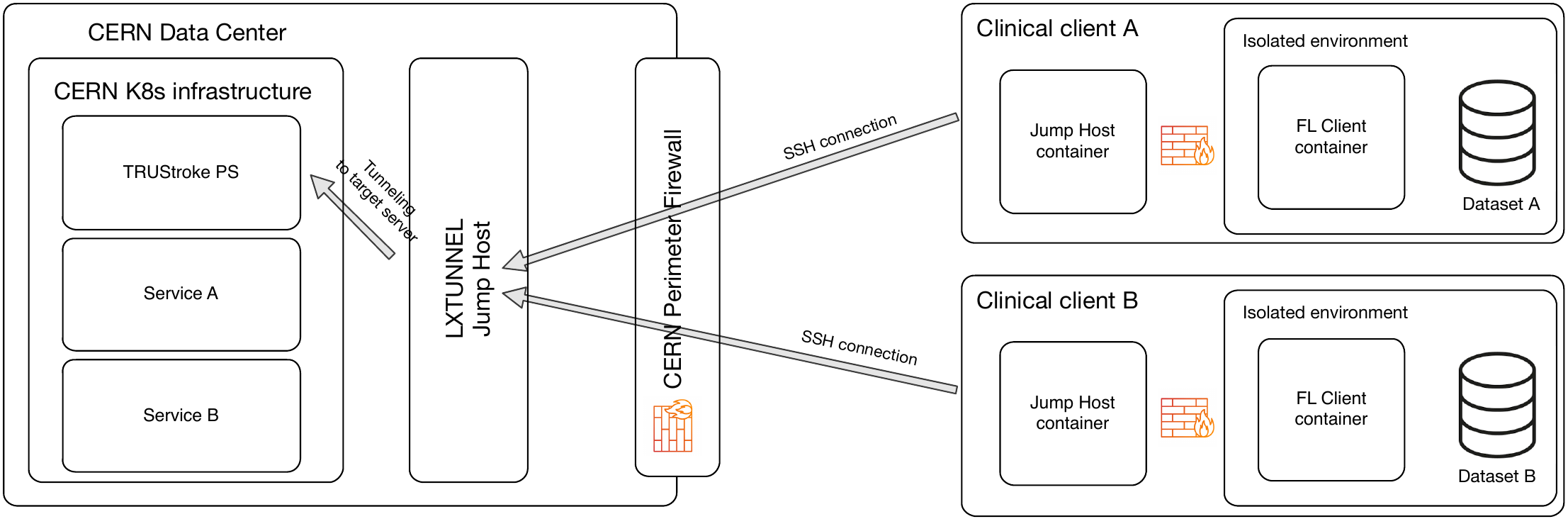}
        \caption{TRUSTroke network architecture: the Parameter Server is hosted within the CERN data center and protected by the security mechanisms implemented thereby. Clinical clients run a Docker-based architecture which (i) decouples sensitive data processing from model parameters communication and (ii) provides portability for easy implementation on different client infrastructures.}
        \label{fig:network_architecture}
\end{figure*}

The TRUSTroke PS is hosted within the CERN Data Centre as a containerized application managed by a Kubernetes (K8s) cluster (Figure \ref{fig:network_architecture}, left). The container networking resources and security aspects are handled by the CERN infrastructure. The key networking elements at CERN include:

\begin{itemize}[leftmargin=*]
\item \textit{Outer Perimeter Firewall:} CERN's network perimeter is protected by a PaloAlto 7080 firewall, offering deep packet inspection and policy-based IP, domain, and URL blocking. It processes incoming and outgoing traffic at 200Gbps, enhancing cybersecurity by blocking unauthorized access and known malicious domains.

\item\textit{Linux Public Logon User Service }(LXTUNNEL): LXTUNNEL enables authenticated access to CERN computing resources and establishes encrypted communication channels via SSH tunnelling, either with  CERN account credentials or Kerberos authentication tokens \cite{steiner1988kerberos}.

\item\textit{Parameter Server}: The TRUSTroke PS runs within a containerized application managed by the K8s cluster, configured to accept connections from clinical nodes. Specific protocols for PS-to-client communication, along with additional security mechanisms for accepting connections from clinical nodes, are detailed in the following sections.
\end{itemize}

\subsection{Client nodes}
Client nodes are represented by clinical institutions which federate their knowledge (local AI models) to create a global model without directly sharing their data. In order to do this, dedicated computing resources (either physical or virtual machines) should be present at the clinical site premises: such machines are required to connect to the PS in order to exchange the local model parameters and contribute to the computation of the global model.
The TRUSTroke project has the ambitious goal of building a FL platform accessible by any clinical institution, which may be characterized by specific IT infrastructure, network policies, etc. To mitigate the differences and complexities arising from the different setups at the clinical nodes and designing an easy-to-use and plug-and-play solution for connecting nodes to the PS, a container-based software architecture based on Docker is proposed for the client nodes. The architecture is shown in Figure \ref{fig:network_architecture}, right: as one can see, the client infrastructure is composed of two main Docker containers named the Jump Host container and the FL client container. 
\begin{enumerate}[leftmargin=*]
\item \textit{FL client container:} this container embeds all the functionalities needed for FL operations, including local model training as well as exchange of parameters with the PS through the Jump Host. Since the FL client container needs access to data for local training, it is critical to isolate it as much as possible from external access. Indeed, such a container is configured with one single network interface, which is assigned a private address in a local network not reachable from outside the clinical node’s internal network. 

\item \textit{Jump Host Container:} this container is responsible for the external connection of the clinical node to the PS and it is configured for such task, including all needed dependencies for easier portability. As an example, the container already contains all functionalities needed for the Kerberos authentication to the CERN servers. The container leverages two different network interfaces, for external and internal communication, respectively.
\begin{enumerate}[leftmargin=*]
\item The internal network interface is solely used for communication with the FL client container. Such interface is assigned an address in the same private local network of the FL client container, which is completely isolated from the Internet.
\item The external network interface is connected to the Internet and is programmed to communicate with CERN’s LXTUNNEL server. The domain lxtunnel.cern.ch is therefore whitelisted for outbound connections on any border firewall at the clinical node’s premises.
The Jump Host uses the external network interface to connect to CERN through a SSH tunnel with port forwarding. All traffic received by the Jump Host on the internal interface from the FL client is therefore encrypted and forwarded to the PS. 
\end{enumerate}
\end{enumerate}

Ultimately, such a container-based design for the client nodes offers two key advantages:
\begin{enumerate}[leftmargin=*]
\item \textit{Consistency and portability:} using Docker containers, which encapsulate everything needed for FL communication and operation, ensure consistent behavior across different environments, and simplifies the task of maintenance and version control. Moreover, Docker containers are highly portable, regardless of the actual underlying infrastructure: this ensures easy migration and adaptation to different environments.
\item \textit{Isolation and security:} the proposed design provides several layers of security both from internal and external threats. Indeed, containers provide process and filesystem isolation, allowing multiple applications to run independently on the same host without interfering with each other. This isolation enhances security and stability by minimizing the impact of one application's issues on others. Moreover, the use of the Jump Host container allows to decouple sensitive data processing for FL operations from model parameters communication through the creation of a so-called Demilitarized Zone (DMZ). The DMZ separates the internal from the external network, creating a buffer zone that limits direct access to internal resources. Additional capabilities such as intrusion detection/prevention system may be deployed within the DMZ to further monitor and regulate traffic flows between the internal and external networks.
\end{enumerate}
\section{Communication protocols: HTTP or MQTT?}\label{sec:comm}
The definition of communication protocols is crucial for the operation of the FL platform within the proposed network architecture. Two main components of communication are identified: the control plane and the data plane. The control plane manages network behavior, including network configuration, client management, and access control, while the data plane handles the forwarding of data packets between clients and the PS in the case of the TRUSTroke FL platform.

For the implementation of the TRUSTroke control and data plane, various communication protocols can be utilized. These protocols fall into two categories based on their communication paradigm: Request/Response and Pub/Sub. In Request/Response protocols, clients send requests to a server, which replies in a one-to-one manner. In Pub/Sub protocols, a message broker facilitates communication between publishers and subscribers, allowing one publisher to reach multiple clients with a single transmission process. HTTP and MQTT are widely recognized standards for the Request/Response and Pub/Sub approaches, respectively.

The following list outlines the communication requirements for the TRUSTroke FL platform:

\begin{itemize}[leftmargin=*]
\item \textit{Asynchronous communication: } the FL process is intrinsically asynchronous, as clients train local models independently from each other and the PS aggregates the received updates according to specific policies. MQTT's asynchronous nature and the publish-subscribe paradigm offer therefore a significant advantage, compared to HTTP's synchronous nature, which may require additional mechanisms and complexity, such as long polling operations, to achieve similar levels of efficiency and real-time communication between the clients and the PS.

\item \textit{One-to-one/one-to-many communication:} for one-to-one communication, both HTTP and MQTT are efficient options for transmitting model parameters from local clients to the parameter server (PS). HTTP request-response model enables direct communication between individual clients and the PS. Similarly, MQTT publish-subscribe model facilitates one-to-one communication, where clients publish updates to specific topics subscribed to by the PS. However, in one-to-many communication scenarios, MQTT holds a natural advantage due to its publish-subscribe paradigm and the presence of a broker. Clients publish updates to a topic, and multiple subscribers (such as the PS and potentially other clients) can receive these updates simultaneously. This simplifies the dissemination of global model parameters to multiple clients, as the PS can publish updates to a topic that all clients subscribe to, ensuring efficient distribution. Conversely, HTTP request-response nature complicates one-to-many communication. To distribute global model parameters to multiple clients via HTTP, additional mechanisms like message queues or specific transmission policies are needed. This complexity and overhead contrast with MQTT built-in support for one-to-many communication.

\item{ \textit{Security:}} the PS should exclusively accept connections from authenticated and authorized clients. CERN's existing security measures already ensure that only clients with valid credentials can connect to the PS. Additionally, the connection occurs over an encrypted channel from the client Jump Host to the CERN Jump Host via the SSH tunnel. To enhance security further, additional measures should be implemented to: (i) regulate the operations that a client can perform on the PS, such as participating in FL tasks or accessing global models, and (ii) encrypt communication between the FL client and the PS end-to-end. For the latter objective, both MQTT and HTTP offer encryption via TLS. Concerning the former, HTTP features an optional authorization mechanism that requests client credentials to access specific resources. In contrast, MQTT offers a natural approach to client authorization through topic access lists. These lists allow for fine-grained access control based on client identifier and permissions.

\item{ \textit{Efficiency and scalability:}} MQTT is indeed designed for efficient communication with low overhead due to its binary-encoded message headers and minimal length. This design makes it particularly suitable for scenarios where bandwidth and resource usage need to be minimized. On the other hand, HTTP relies on textual headers, which can introduce significant overhead, especially for small payloads. However, both protocols can handle large payloads effectively, as they do not impose specific upper limits on message size.
In terms of scalability, MQTT message brokers are naturally designed to handle a massive number of clients, potentially in the order of millions. This scalability is inherent to the publish-subscribe architecture of MQTT. In contrast, HTTP web servers may require specific implementations and optimizations to handle a large number of concurrent connections efficiently.
Overall, while both MQTT and HTTP are capable of exchanging large payloads, MQTT's efficiency and scalability make it well-suited for scenarios requiring lightweight communication and handling a large number of clients.
\end{itemize}

Table \ref{tab:mqttvshttp} summarizes the main differences between HTTP and MQTT, demonstrating that the Pub/Sub approach aligns better with the requirements of the TRUSTroke FL platform. Therefore, the MQTT protocol has been selected as the communication protocol to implement both the Control and Data plane for the initial release of the TRUSTroke FL platform. It's important to note that the MQTT protocol requires an MQTT broker to act as an intermediary between publishers and subscribers. This broker should ideally be located as close as possible to the PS to minimize communication latency. Therefore, the TRUSTroke FL platform's MQTT broker can be deployed in the same virtualized instance and local network as the PS, leveraging CERN's abundant computational and communication resources.

\begin{table}[t]
    \centering
    \caption{Communication protocols comparison}
    \begin{tabular}{|l|c|c|}
    \hline
         \textbf{Feature} & \textbf{HTTP} & \textbf{MQTT} \\
         \hline
    Communication paradigm     & Synchronous & Asynchronous\\
    Authentication/authorization & Managed ad hoc & Provided by broker \\
    Message encryption & Supported (TLS)  & Supported (TLS)\\
    One-to-many communication     &  Managed ad hoc & Provided by broker\\
    Message overhead     & 4-8KB & 2B min, 14B max\\
    Max. payload size     & No limit & 256MB\\
    Scalability     & Managed ad hoc & Provided by broker \\
    \hline
    \end{tabular}
    \label{tab:mqttvshttp}
\end{table}

\section{TRUSTroke Control and Data plane}\label{sec:plane}
\subsection{Control plane}
In order to properly design the control plane for the TRUSTroke FL platform, a requirement analysis of the functionalities of a generic Federated Learning system has been conducted. This preliminary study also involved the analysis of publicly available FL platforms such as Flower \cite{beutel2020flower} and OpenFL \cite{reina2021openfl}. 
The following list details the control plane functionalities that have been identified and how they are efficiently mapped to MQTT message exchange:
\begin{itemize}[leftmargin=*]
\item \textit{Client authentication and authorization:} the PS will manage several
concurrent learning tasks, named Clinical End Points (CEPs).
Each CEP responds to a different objective and possibly
require different data to be trained. Moreover, different sets
of clinical clients (i.e., multiple federations) may be active at
the same time, cooperating on specific CEPs. It follows that a
client may be willing to join only specific federations
and CEPs managed by the PS. A client's request to join undergoes a screening process which assesses
the clinical client characteristics (security features, clinical
features needed for the learning model, etc.). If the screening process results in a positive assessment, the clinical node receives a set of credentials, namely: (i) a CERN account to be able to produce a Kerberos token needed for SSH access to the CERN infrastructure, (ii) cryptographic material needed for properly authenticating with the MQTT broker (either username and password or a digital certificate), together with a unique MQTT \texttt{client\_id} and (iii) the set of topics it is authorized to publish and subscribe to (detailed later). Note that both authentication and authorization mechanisms are provided by most MQTT brokers and do not need to be implemented ad hoc in the FL platform.





where client\_id is the identifier assigned to the clinical client after the screening process.

\item \textit{Client joining/leaving:} Clients may need to join or leave federations dynamically based on their needs or the availability of tasks. Joining a federation allows a client to participate in collaborative learning, while leaving a federation may occur if the client no longer wishes to contribute or if its resources become unavailable. After successful connection to the PS using the personal credentials, a client willing to join a specific federation and CEP is simply required to perform a subscribe operation on the following topic (named T1 for brevity):

\begin{equation}
    \texttt{fed\_id/CEP\_id/job\_request}
    \label{topic:job_req}
\end{equation}

where \texttt{federation\_id} and \texttt{CEP\_id} are the identifiers of the federation and CEP the clients is willing to join, and for which it is authorized. Similarly, a client willing to leave the FL process may simply unsubscribe from the aforementioned topic. 

\item \textit{Runtime operations:} The PS and the client nodes use two specific MQTT topics for exchanging the model parameters. In particular, the PS uses the topic \eqref{topic:job_req} to publish the initial model template and the computed global models to the client. Conversely, the clients use the following topic for updating the local parameters to the PS:
\begin{center}
\texttt{fed\_id/CEP\_id/job\_replies/client\_id}
\end{center}
To receive such updates, the PS subscribes to:
\begin{center}
    \texttt{fed\_id/CEP\_id/job\_replies/\#}
\end{center}
The use of the hash wildcard allows to efficiently receive updates from all connected clients in a single operation.

\item \textit{Global model versioning:} This action involves managing the versioning of the global model within the federated learning network. As the model evolves through training iterations and updates from participating clients, versioning ensures consistency and integrity across the network, allowing for proper synchronization and coordination of learning efforts. At any point in time, a client with authorized access may retrieve the list of global models available on the PS for a specific federation and CEP. Such an operation is performed by a client with a specific publish message on the topic, where the PS is subscribed:
\begin{center}
\texttt{fed\_id/CEP\_id/model\_request/client\_id}
\end{center}
The PS uses the following topic for replying:
\begin{center}
\texttt{fed\_id/CEP\_id/model\_reply/client\_id}
\end{center}
The two topics create a bidirectional communication channel between any client and the PS: the channel is used by (i) a client, to request the list of available global models on the PS for the specific CEP, possibly including information such as which other clients participated in the global model training, (ii) a client, to request the download of a specific model version, (iii) the PS to transmit the requested global model to a client.
\end{itemize}

\subsection{Data plane}
For what concerns the data plane, the MQTT protocol already offers the main features needed for an effortless implementation. In particular:
\begin{itemize}[leftmargin=*]
\item \textit{Custom payload formats:} MQTT payloads can be formatted in any custom format or structure based on the specific needs of the application. This allows for flexibility in designing message formats that best suit the requirements of the application. In this specific case, MQTT payloads are used for transmitting both control information as well as model parameters. We suggest the use of compressed Protocol Buffer (protobuf) for the serialization of the message payloads, rather than the transmission of uncompressed text-based JSON payloads. This allows for smaller message sizes, faster serialization and deserialization, and backward and forward compatibility.

\item \textit{Quality of service:} MQTT offers three incremental levels of QoS, which can be used as an additional error control method over the one already offered by the TCP protocol. For the initial release of the TRUSTroke FL platform, given the small size of the network scenario (one PS and three clinical clients) the minimum level of QoS will be used and error control will be devoted to the underlying TCP protocol. The QoS level may be easily increased to level 1 or 2, which requires acknowledgements also at the application layer, in case the error control provided by TCP is not sufficient.

\item \textit{Encryption:} all communications between clients and the PS will be automatically encrypted end-to-end by TLS, which is naturally supported by MQTT. TLS ensures that the data exchanged between MQTT clients and brokers is encrypted and authenticated, thus enhancing the security of the data plane.
\end{itemize}
\begin{table*}
    \caption{Security threats associated with the proposed architecture}
    \centering
    \begin{tabular}{|p{30mm}|p{60mm}|p{30mm}|p{45mm}|}
    \hline
       Threat  & Description  & Attacked component & Mitigation \\
       \hline
    Data Breach/Leakage & Unauthorized access to patient data poses a significant risk if client containers are compromised, potentially leading to severe privacy breaches and reputational damage to the TRUSTroke project. & Clinical client & The FL client and local dataset are isolated from external connections by design.\\
    \hline
    Data Interception and Tampering & Unauthorized interception or alteration of model parameters and training data during transmission, which could impact the reliability and performance of global and local models. & Communication infrastructure & Use of MQTT data plane with TLS encryption\\
    \hline 
    Unauthorized Access & The risk of unauthorized individuals gaining access to client or server systems is high, especially if security measures outside the Federated Learning (FL) Client Container are inadequate. This could lead to data theft and system compromise, including access to the parameter server if security isn't sufficient. & Parameter Server,  Clinical client & Access to PS is regulated by SSH tunneling, Kerberos authentication and MQTT5 enhanced authentication. The client is made accessible only from within the clinical infrastructure network.\\
    \hline
    Denial of Service Attacks & Overloading the PS MQTT broker to disrupt the training process. & Parameter server & Adopt MQTT5\cite{morelli2021attacks}, which reduces the risk of DoS due to message queue filling at the broker.  \\
    \hline
    Docker Configuration and Patch Management Failures & Improper Docker system configuration or failure to apply security patches timely may expose the system to known vulnerabilities. & Clinical client  & Follow Open Web Application Security Project (OWASP) guidelines for secure Docker configuration\cite{ahamed2021security}\\
    \hline
    \end{tabular}
    \label{tab:security}
\end{table*}

\section{Security considerations}\label{sec:security}
Security measures are introduced for protecting data, preventing exploitation and mitigating risks, overall fostering general confidence and trust in the proposed solution. Table \ref{tab:security} provides a brief description of the threats and risks associated with each component (Parameter Server, Clinical client or Communication infrastructure) of the proposed architecture, as well as the adopted mitigation technique. As one can see, the main identified threats are well mitigated by the design choices taken within the proposed system. Following the threat identification step, a series of security recommendation may be compiled for each component of the proposed architecture.

\subsection{Parameter Server} In the context of server-side security and operational specifications, the requirements primarily focus on authorization processes, access control for services such as the parameter server and MQTT broker, and the education of intern personnel on security protocols. It is pertinent to note that the network infrastructure at CERN already complies with these established requirements, furthermore highlighting CERN's appropriateness as serving for a public PS for FL processes.

\subsection{Communication infrastructure} The main requirements for the communication infrastructure are linked to authentication and communication integrity and confidentiality. The use of SSH tunnels and authentication with Kerberos together with MQTT authentication primitives at the PS provides authentication. Similarly, using TLS with MQTT allows for message integrity and confidentiality. 

\subsection{Clinical clients} The clinical clients are probably the most vulnerable component of the proposed architecture, given that (i) they host sensitive data and (ii) each clinical client is based on a different IT infrastructure, possibly implementing different security mechanisms. The proposed design for the client node allows to mitigate many of the associated threats, although others actions should be taken at the clinical infrastructure to increase security. As an example, it is necessary to design an access control system that limits direct access to the machine on which the containers run. During the design of such an access control system, the principle of least privileges (PoLP) should be applied, i.e., users should be given the minimum access levels or permissions they need to perform their job functions. Furthermore, it is important to conduct regular penetration testing and vulnerability assessments on the client's side. A robust patch management system should be implemented to ensure timely application of security updates. Then, detailed logging and monitoring of all system activities and data access are required for improved threat detection and response.

\section{Conclusion}\label{sec:concl}
This paper introduced the design solution for the network architecture and the communication protocols to be used within the TRUSTroke FL platform. The network design is based on a centralized architecture, with a Parameter Server hosted at CERN’s premises and the clients running in the clinical nodes’ infrastructures. A Docker-based client design is proposed, which decouples the communication and data access/processing part for increased privacy and security. The paper also introduces requirements for the control and data plane, selecting the MQTT protocol as reference communication protocol for operating the FL platform and defining a specific control plane to support such operations. As a final contribution, the paper reports on security aspects of the proposed design. Future research direction will include the implementation and testing of the proposed FL platform for stroke prediction as well as its comparison with existing frameworks for general FL operations. 

\balance



\bibliographystyle{IEEEtran}
\bibliography{IEEEabrv,bibliography}
\vspace{12pt}

\end{document}